\documentclass[10pt,twocolumn,letterpaper]{article}
\usepackage{cvpr}
\usepackage{times}
\usepackage{epsfig}
\usepackage{graphicx}
\usepackage{amsthm,amssymb,mathtools}
\usepackage{subfigure}
\usepackage{CJK}
\usepackage{multirow}
\usepackage{booktabs}
\usepackage{array}
\usepackage{color}
\usepackage{algorithm}
\usepackage{algorithmic}
\usepackage{enumitem}

\newcolumntype{L}[1]{>{\raggedright\let\newline\\\arraybackslash\hspace{0pt}}m{#1}}
\newcolumntype{C}[1]{>{\centering\let\newline  \\\arraybackslash\hspace{0pt}}m{#1}}
\newcolumntype{R}[1]{>{\raggedleft\let\newline \\\arraybackslash\hspace{0pt}}m{#1}}

\newtheorem{theorem}{Theorem}[section]

\newtheorem{proposition}[theorem]{Proposition}

\newtheorem{remark}{Remark}[section]

\usepackage[pagebackref=true,breaklinks=true,letterpaper=true,colorlinks,bookmarks=false]{hyperref}
 \cvprfinalcopy 
\def\cvprPaperID{6541} 
\ifcvprfinal\fi

\begin{document}
\title{Non-local Meets Global: An Integrated Paradigm for Hyperspectral Denoising}

\author{
Wei~He$^{1}$, Quanming Yao$^{2}$\thanks{Q. Yao is the corresponding author,
	and the work is done he is working in 4Paradigm. Inc.} , Chao~Li$^1$, Naoto~Yokoya$^1$\thanks{This work was supported by the Japan Society for the Promotion of Science (KAKENHI 18K18067).} , Qibin~Zhao$^1$\\
$^1$RIKEN AIP\ \ \ $^2$HKUST \\
\tt\small \{wei.he;Chao.Li;naoto.yokoya;qibin.Zhao\}@riken.jp, qyaoaa@connect.ust.hk
}

\maketitle

\begin{abstract}
Non-local low-rank tensor approximation has been developed as a state-of-the-art method for hyperspectral image
(HSI) denoising.
Unfortunately, while their denoising performance benefits little from more spectral bands, the running time of these methods significantly increases. In this paper, we claim that the HSI lies in a global spectral low-rank subspace, and the spectral subspaces of each full band patch groups should lie in this global low-rank subspace. This motivates us to propose a unified spatial-spectral paradigm for HSI denoising. As the new model is hard to optimize, An efficient algorithm motivated by alternating minimization is developed. This is done by first learning a low-dimensional orthogonal basis and the related reduced image from the noisy HSI. Then, the non-local low-rank denoising and iterative regularization are developed to refine the reduced image and orthogonal basis, respectively. Finally, the experiments on synthetic and both real datasets demonstrate the superiority against the state-of-the-art HSI denoising methods.
\end{abstract}

\vspace{-10px}
\section{Introduction}
Recent decades have witnessed the development of hyperspectral imaging techniques~\cite{HarvardCVPR2011,CAVETIP2010,Green1998}. The hyperspectral imaging system is able to cover the wavelength region from 0.4 to 2.5$\mu m$ at a nominal spectral resolution of 10 nm. With the wealth of available spectral information, hyperspectral images (HSI) have the high spectral diagnosis ability to distinguish precise details even between the similar materials~\cite{Bioucas2012jstars,Shaw2003}, providing the potential advantages of application in remote sensing~\cite{Stein2002,facehyperTIP2015}, medical diagnosis~\cite{Medicalhyper2014}, face recognition~\cite{HyperfaceTPAMI2003,facehyperTIP2015}, quality control~\cite{GENDRIN2008} and so on. Due to instrumental noise, HSI is often corrupted by Gaussian noise, which significantly influences the subsequent applications. As a preprocessing, HSI denoising is a fundamental step prior to HSI exploitation~\cite{changyiTIP2015,He2014TGRS,weiweiICCV015}.

For HSI denoising,
the spatial non-local similarity and global spectral correlation are the two most important properties.
The spatial non-local similarity suggests that similar patches inside a HSI can be grouped and denoised together.
The related methods
\cite{baiJSTAR2018,chang2017hyper,lishutaoCVPR2017,DongCS2014,Dong2015ICCV,CVPR2014Meng,xie2017kronecker,zhang2018nonlocal} denoise the HSIs via group matching of \textit{full band patches} (FBPs, stacked by patches at the same location of HSI over all bands) and low-rank denoising of each non-local FBP group (NLFBPG).
These methods have achieved state-of-the-art performance.
However, they still face a crucial problem.
For HSIs, the higher spectral dimension means the higher discriminant ability~\cite{Bioucas2012jstars},
thus more spectrums are desired.
As the spectral number increases, the size of NLFBPG also becomes larger,
leading to significantly more computations for the
subsequent low-rank matrix/tensor approximations.

The HSIs have strong spectral correlation,
which is modeled as low-rank property \cite{BioucasTGRS2008,DuqianTGRS1999, Bourennane2008,he2018noise,He2014TGRS} and have also been widely adopted to the HSI denoising.
However, due to the lack of spatial regularization, only spectral low-rank regularization cannot remove the noise efficiently.
One promising improvement is to project the original noisy HSI onto the low-dimensional spectral subspace,
and denoise the projected HSI via spatial based methods~\cite{Chen2011TGRS,rasti2014hyperspectral,zhuang2018fast}.
Unfortunately,
these two-stage methods are significantly influenced by the quality of projection and the efficiency of spatial denoising.
All of them fail to capture a clean projection matrix,
which makes the restored HSI still be noisy.


\begin{figure*}[t]
	\centering
	\vspace{-20px}
	\includegraphics[width= 0.95 \linewidth]{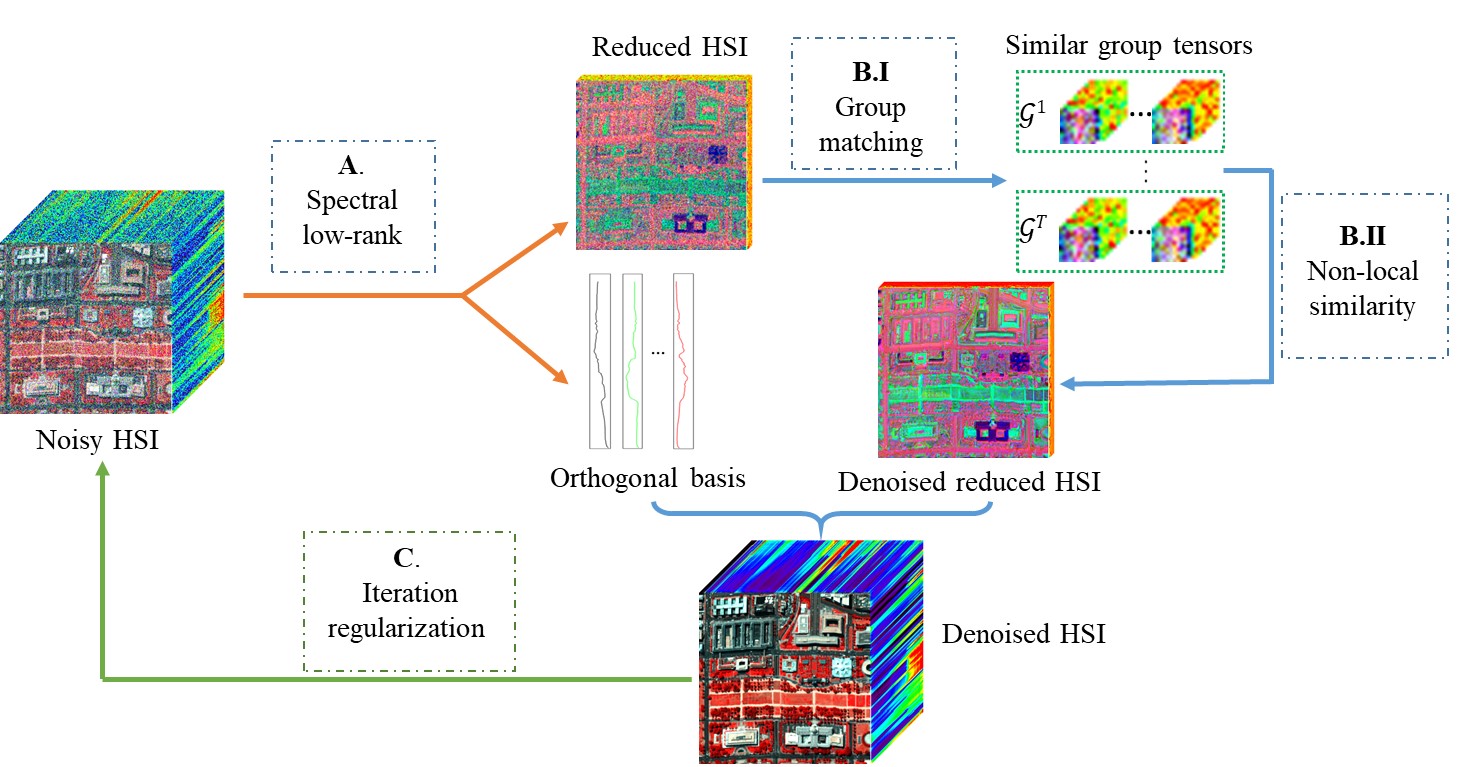}
	\caption{Flowchart of the proposed method. It includes three stages: A. spectral low-rank denoising, B. non-local low-rank denoising and C. iteration regularization. B consists of two steps including group matching and non-local low-rank approximation.}
	\label{fig:flow_m1}
	\vspace{-10px}
\end{figure*}

To alleviate the aforementioned problems,
this paper introduces
a unified HSI denoising paradigm to integrate the spatial non-local similarity and global spectral low-rank property simultaneously.
We start from the point that the HSI should lie in a low-dimensional spectral subspace, which has been widely accepted in hyperspectral imaging~\cite{FuCVPR2016}, compressive sensing~\cite{MengTIP2016,weiwei2015CVPR}, unmixing~\cite{Bioucas2012jstars} and dimension reduction~\cite{BioucasTGRS2008} tasks. Inspired by this fact, the whole NLFBPGs should also lie in a common low-dimensional spectral subspace.
Thus, we first learn a global spectral low-rank orthogonal basis,
and subsequently exploit the spatial non-local similarity of projected HSI on this basis.
The computational cost of non-local processing in our paradigm will almost keep the same with more spectral bands,
and the global spectral low-rank property will also be enhanced.
The contributions are summarized as follows:
\begin{itemize}[leftmargin = 10pt]
\item We propose a new paradigm for HSI denoising, which can \textit{jointly learn} and \textit{iteratively update} the orthogonal basis matrix and reduced image.
This is also the first work successfully combines the power of existing spatial and spectral denoising methods;


\item The resulting new model for image denoising is hard to optimize,
as it involves with both complex constraint
(from spectral denoising) and regularization
(from spatial denoising).
We further propose an efficient and iterative algorithm for optimization,
which is inspired by alternating minimization;

\item Finally,
the proposed method is not only the best compared with other state-of-the-art methods in simulated experiment,
where Gaussian noise are added manually;
but also achieves the most appealing recovered images for real datasets.
\end{itemize}

\noindent
\textbf{Notations.}
We follow the tensor notation in \cite{Kolda2009},
the tensor and matrix are represented as Euler script letters, \textit{i.e.} $\mathcal{X}$ and boldface capital letter, \textit{i.e.} $\mathbf{A}$, respectively.
For a $N$-order tensor $\mathcal{X}\in\mathbb{R}^{I_1\times I_2\times\cdots \times I_N}$,
the mode-$n$ unfolding operator is denoted as $\mathbf{X}_{(n)}\in\mathbb{R}^{I_n \times  {I_1 \cdots I_{n-1} I_{n+1} \cdots I_N}}$.
We have $\text{fold}_n(\mathbf{X}_{(n)})=\mathcal{X}$, in which $\text{fold}_n$ is the inverse operator of unfolding operator. The Frobenius norm of $\mathcal{X}$ is defined by $\left \| \mathcal{X} \right \|_F= (\sum_{i_1}\sum_{i_2}\cdots\sum_{i_N}x_{i_1 i_2\ldots i_N}^2)^{0.5}$.
The mode-$n$ product of a tensor $\mathcal{X}\in\mathbb{R}^{I_1\times I_2\times\cdots \times I_N}$ and a matrix
$\mathbf{A}\in\mathbb{R}^{J_n\times I_n}$ is defined as $\mathcal{Y} = \mathcal{X} \times_n
\mathbf{A}$, where $\mathcal{Y}\in\mathbb{R}^{I_1\times I_2\times\cdots \times J_n}$ and $\mathcal{X}\times_n \mathbf{A} = \text{fold}_n(\mathbf{A} \mathbf{X}_{(n)})$.

\vspace{-3px}
\section{Related work}

Since denoising is an ill-posed problem, proper regulations based on the HSI prior knowledge is necessary~\cite{fu2017adaptive,wei2017structured}.
The mainstream of HSI denoising methods can be grouped into two categories:
spatial non-local based methods and spectral low-rank based methods.

\subsection{Spatial: Non-local similarity}

HSIs illustrate the strong spatial non-local similarity. After the non-local low-rank modeling was first introduced to HSI denoising in~\cite{CVPR2014Meng}, the flowchart of the non-local based methods become fixed: FBPs grouping and low-rank tensor approximation. Almost all the researchers focused on the low-rank tensor modeling of NLFBPGs, such as tucker decomposition~\cite{CVPR2014Meng}, sparsity regularized tucker decomposition~\cite{xie2017kronecker}, Laplacian scale mixture low-rank modeling~\cite{Dong2015ICCV}, and weighted low-rank tensor recovery \cite{chang2017weighted} to exploit the spatial non-local similarity and spectral low-rank property simultaneously. However, with the increase of spectral number, the computational burden also increases significantly, impeding the application of these methods to the real high-spectrum HSIs.

Chang et.al~\cite{chang2017hyper} claimed that the spectral low-rank property of NLFBPGs is weak and proposed a unidirectional low-rank tensor recovery to explore the non-local similarity. It saved much computational burden and achieved the state-of-the-art performance in the HSI denoising. This reflects the fact that previous non-local low-rank based methods have not yet efficiently utilized the spectral low-rank property. How to balance the importance between spectral low-rank and spatial non-local similarity still remains a problem.
\vspace{-3px}
\subsection{Spectral: Global low-rank property}
\label{sec:rel:global}

The global spectral low-rank property of HSI has been widely accepted and applied to the subsequent applications~\cite{BioucasTGRS2008,DuqianTGRS1999}.
As pointed out in~\cite{BioucasTGRS2008},
the intrinsic dimension of the spectral subspace is far less than the spectral dimension of the original image.
By vectorizing each band of the HSI and reshaping the original 3-D HSI into a 2-D matrix, various low-rank approximation methods such as principal components analysis (PCA)~\cite{DuqianTGRS1999,yao2018scalable},
robust PCA~\cite{chen2017denoising,xie2016hyperspectral,He2014TGRS},
low-rank matrix factorization~\cite{MengTIP2016,QianyuntaoTGRS2015} have been directly adopted to denoise the HSI.
However, these methods only explore the spectral prior of the HSI, ignoring the spatial prior information. Instantly, many conventional spatial regularizers such as total variation~\cite{HE2016TGRS}, low-rank tensor regularization \cite{li2015hyperspectral,renard2008denoising} are adopted to explore the spatial prior of HSI combined with spectral low-rank property.

A remedy is a two-stage method combining the spatial regularizer and spectral low-rank property together. This is done by firstly mapping the original HSI into the low-dimensional spectral subspace, and then denoise the mapped image via existing spatial denoising methods, e.g., wavelets \cite{Chen2011TGRS,rasti2014hyperspectral},
BM3D~\cite{zhuang2018fast} and HOSVD~\cite{zhuang2017hyperspectral}.
These two-stage methods provide a new sight to denoise the HSI in the transferred spectral space,
which is very fast.
However,
these methods do not \textit{iteratively} refine the subspace
and thus fail to combine the best of both worlds,
and the extracted subspace is still corrupted by the noise.

\section{The Proposed Approach}
In this section, we propose a unified HSI denoising paradigm to integrate
spatial non-local similarity
and global spectral low-rank property. We first learn a low-dimensional orthogonal basis and the related reduced image from the noisy HSI. Then the reduced image and the orthogonal basis are updated by spatial non-local denoising and iteration regularization, respectively.
The overview of the proposed paradigm is in Figure~\ref{fig:flow_m1}.

\subsection{Unified spatial-spectral paradigm}
\label{sec:framework}

Assuming that the clean HSI $\mathcal{X} \in \mathbb{R}^{M \times N \times B}$ is corrupted by
the additive Gaussian noise $\mathcal{N}$ (with zero mean and variance $\sigma_0^2$),
then the noisy HSI $\mathcal{Y}$ is generated by
\begin{equation}
\mathcal{Y} = \mathcal{X} + \mathcal{N}.
\label{eq:basic}
\end{equation}

First, to capture the spectral low-rank property in Section~\ref{sec:rel:global},
we are motivated to use a low-rank representation of the clean HSI $\mathcal{X}$,
\textit{i.e.}
$\mathcal{X} = \mathcal{M} \times_3 \mathbf{A}$,
where $K \ll B$, $\mathbf{A}\in \mathbb{R}^{B \times K}$ is an orthogonal basis matrix capturing the common subspace of different spectrum,
and $\mathcal{M}\in \mathbb{R}^{M \times N \times K}$ is the reduced image.
Second,
to utilize the spatial low-rank property,
we add a non-local low-rank regularizer $\| \cdot \|_{\text{NL}}$ on the reduced image $\mathcal{M}$.
As a result, the proposed non-local meets global (NGmeet) denoising paradigm is presented as
\begin{align}
\{\mathcal{M}_*, \mathbf{A}_*\}
= \arg\min_{\mathcal{M}, \mathbf{A}}
& \frac{1}{2}
\Vert \mathcal{Y}
\!\times_3\!
\mathbf{A}^{\top} - \mathcal{M} \Vert_F^2+ \mu{\Vert \mathcal{M} \Vert_{\text{NL}}},
\notag
\\
&
\text{s.t.\;}
\mathbf{A}^{\top} \mathbf{A} = \mathbf{I},
\label{eq:overall}
\end{align}
where $\mu$ controls the contribution of spatial non-local regularization, the basis matrix $\mathbf{A}$ is required to be orthogonal,
and the clean HSI is recovered by $\mathcal{X} = \mathcal{M}_* \times_3 \mathbf{A}_*$.

The objective \eqref{eq:overall} is very hard to optimize,
due to both the orthogonal constraint on $\mathbf{A}$ and
complex regularization on $\mathcal{M}$.
An algorithm based on alternating minimization to approximately solve the objective function is proposed in
Section~\ref{sec:opt}.

\begin{remark}
	The orthogonal constraint $\mathbf{A}^{\top} \mathbf{A} = \mathbf{I}$ is very important here.
	First,
	it encourages the representation held
	in $\mathbf{A}$ to be more distinguish with each other.
	This helps to keep noise out of $\mathbf{A}$ and
	further allows a closed-form solution for computing $\mathbf{A}$ (Section~\ref{sec:opt:spectral}).
	Besides,
	it preserves the distribution of noise,
	which allows us to estimate the remained noise-level in reduced image and
	reuse state-of-the-art Gaussian based non-local method for spatial denoising (Section~\ref{sec:opt_spatial}).
\end{remark}

However,
before going to optimization details,
we first look into \eqref{eq:overall},
and see the insights why the proposed method can beat
all previous spectral low-rank methods~\cite{Chen2011TGRS,zhuang2018fast}.

\begin{figure}[t]
	\centering
	\includegraphics[width= 0.9 \linewidth]{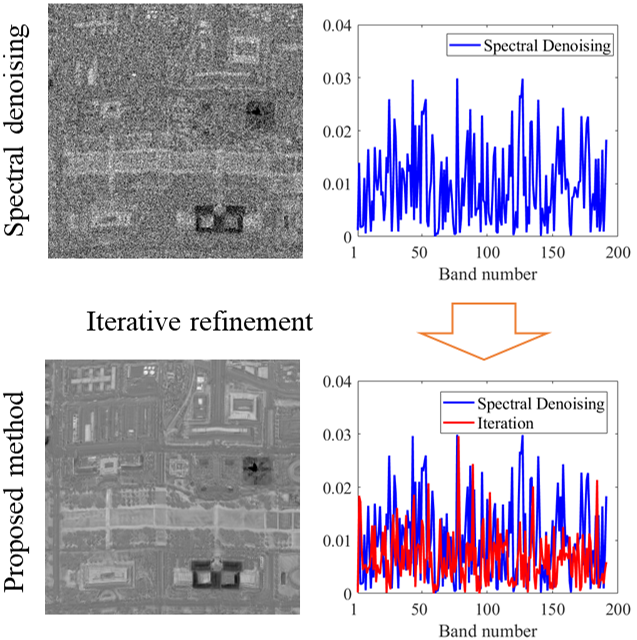}
	\caption{
		The first row displays the coefficient image $\bar{\mathcal{M}}(:,:,4)$ and the absolute difference signature between $\mathbf{A}_1(:,4)$ and the reference.
		The second row displays the refined coefficient image and the absolute difference signature between refined one and the reference. The test dataset is WDC with noise variance 50.
	}
	\vspace{-10px}
	\label{fig:Motivation1}
\end{figure}

\vspace{-5px}
\subsubsection{Necessity of iterative refinement}
Recall that,
in~\eqref{eq:overall},
the first item tries to exploit the spectral low-rank property and decompose the noisy $\mathcal{Y}$ into the coarse spectral
low-rank orthogonal basis $\mathbf{A}$ and reduced image $\mathcal{M}$.
Specifically,
$i$-th column of $\mathbf{A}$,
denoted as $\mathbf{A}(:,i)$,
is regarded as the $i$-th signature of HSI,
and the corresponding coefficient image $\mathcal{M}(:,:,i)$ is regarded as the abundance map.

Previous methods are mostly two-stage ones,
they do not iterative refine the orthogonal basis matrix they found,
\textit{e.g.} FastHyDe~\cite{zhang2018nonlocal}.
However,
we model the spatial and spectral low-rank properties simultaneously,
which enables iterative refinement of the orthogonal basis matrix $\mathbf{A}$.
To demonstrate the necessity of iterative refinement,
we calculated the orthogonal basis $\mathbf{A}_1$ and reduced image $\bar{\mathcal{M}}$ from noisy WDC with noise variance 50. The reference ${\mathbf{A}}$ and ${\mathcal{M}}$ are from the original clean WDC. Figure~\ref{fig:Motivation1} presents the
comparison on signatures and the corresponding coefficient image before and after our refinement.
From the figure,
it can be observed that the orthogonal basis atom $\mathbf{A}_1(:,4)$ and reduced image $\bar{\mathcal{M}}$ obtained by the spectral denoising method are still suffering from the noise, while the proposed method produces much cleaner  signatures and  coefficient images.

\subsection{Efficient optimization}
\label{sec:opt}
As discussed in Section~\ref{sec:framework},
the objective \eqref{eq:overall} is very hard to optimize.
In this section,
we are motivated to use alternating minimization for optimization
(Algorithm~\ref{alg:NSLR}). $\mathcal{Y}_i$, $\mathcal{X}_i$ stand for the input noisy image and output denoised image of the $i$-th iteration, respectively.
As will be shown in the sequel,
Algorithm~\ref{alg:NSLR} tries to find a closed-form solution for $\mathbf{A}$ (step~3)
and reuses state-of-the-art spatial denosing method for computing $\| \cdot \|_{\text{NL}}$ (steps~4-6),
which together make the algorithm very efficient.
Besides,
as $\mathbf{A}$ will be refined during the iteration,
iterative regularization~\cite{dong2013nonlocal}
is adopt to boost the denosing performance (step~7).

\begin{algorithm}[ht]
\caption{Non-local Meets Global(NGmeet)}
\label{alg:NSLR}
\begin{algorithmic}[1]
	\REQUIRE Noisy image $\mathcal{Y}$, noise variance $\sigma_0^2$
	\STATE $\mathcal{X}_1=\mathcal{Y}_1=\mathcal{Y}$, estimating $K$ using HySime~\cite{BioucasTGRS2008};
	\FOR{$i = 1, 2, 3, \cdots iter$}
	\STATE {A). \textit{Spectral low-rank denoising}}:
	\\
	 {Estimate orthogonal basis matrix $\mathbf{A}_i$ and reduced image $\bar{\mathcal{M}}_i$ via SVD on $\mathcal{Y}_i$;}
	\STATE {B). \textit{Non-local reduced image $\bar{\mathcal{M}}_i$ denoising}:}
	\\
	{-B.I) Obtain the set of tensors $\left\{\mathcal{G}^j\right\}$
		for $\bar{\mathcal{M}}_i$ via $k$-NN search for each reference patch;}
	\STATE {-B.II) Denoise $\{ \mathcal{G}^j \}$ via Low-rank approximation and obtain $\{\mathcal{M}_i^j\}$;}
	\STATE {-B.III) Reconstruct the cubes $\{ \mathcal{M}_i^j \}$ to image $\mathcal{M}_i$,
		and obtain the denoised HSI $\mathcal{X}_i=\mathcal{M}_i\times_3\mathbf{A}_i$;}
	\STATE (C). \textit{Iterative regularization:} \\
	$\mathcal{Y}_{i+1}=\lambda\mathcal{X}_i+(1-\lambda)\mathcal{Y}$, $K=K+\delta \times i$;
	\ENDFOR
	
	\RETURN  Denoised image $\mathcal{X}_i$;
\end{algorithmic}
\end{algorithm}

\vspace{-5px}
\subsubsection{Spectral denoising via $\mathbf{A}$}
\label{sec:opt:spectral}

In this stage,
we identify the orthogonal basis matrix $\mathbf{A}$ with
the given $\mathcal{M}_i$ and $\mathcal{Y}_i$ from \eqref{eq:overall},
which leads to
\begin{align}
\arg\min_{\mathbf{A}^{\top} \mathbf{A} = \mathbf{I}}
\frac{1}{2} \Vert \mathcal{Y}_i \times_3 \mathbf{A}^{\top} - \mathcal{M}_i \Vert_F^2.
\label{eq:sub:a}
\end{align}
However,
this problem is hard without simple closed-form solution.
Instead,
since $\mathcal{Y}_i$ is obtained from iterative regularization,
of which the noisy-level is decreased.
Thus,
we proposed to relax \eqref{eq:sub:a} as
\begin{equation}
\{ \bar{\mathcal{M}}_i, \mathbf{A}_i \}
= \arg \min_{\mathcal{M},\mathbf{A}^{\top} \!\! \mathbf{A}=\mathbf{I}}
\frac{1}{2}
\| \mathcal{Y}_i - \mathcal{M}\times_3\mathbf{A} \|_F^2,
\label{eq:diclearning}
\end{equation}
which has the closed-form solution (Proposition~\ref{pr:svd}).
Thus,
only a SVD on the folding matrix of $(\mathcal{Y}_i)_{(3)}$ is required,
which can be efficiently computed.

\begin{proposition} \label{pr:svd}
Let $(\mathcal{Y}_i)_{(3)}=\mathbf{U}\mathbf{S} \mathbf{V}^\top$ be the rank-$K$ SVD of $(\mathcal{Y}_i)_{(3)}$.
The solution to \eqref{eq:diclearning}
is given by the close-form as $\mathbf{A}_i = \mathbf{V}$ and $\bar{\mathcal{M}_i} = \text{\normalfont fold}_3(\mathbf{U} \mathbf{S})$.
\end{proposition}

\vspace{-5px}
\subsubsection{Spatial denoising via $\mathcal{M}$}
\label{sec:opt_spatial}

Note that we have $\bar{\mathcal{M}}_i = \mathcal{Y}_i \times_3 \mathbf{A}_i^\top$
from Section~\ref{sec:opt:spectral}.
Using $\bar{\mathcal{M}}_i$ in \eqref{eq:overall},
the objective in this stage becomes:
\vspace{-3px}
\begin{equation}
\mathcal{M}_i = \arg\min_{\mathcal{M}}
 \frac{1}{2}
\Vert \bar{\mathcal{M}}_i - \mathcal{M} \Vert_F^2+ \mu{\Vert \mathcal{M} \Vert_{\text{NL}}},
\label{eq:opt_spatial}
\end{equation}
where $\Vert \cdot \Vert_{\text{NL}}$ is a non-local denoising regularizer.
Formulation \eqref{eq:opt_spatial} appears in many denoising models,
\textit{e.g.} WNNM~\cite{gu2014weighted},
TV \cite{HE2016TGRS}, wavelets \cite{Chen2011TGRS,rasti2014hyperspectral} and CNN \cite{chang2018hsi}.
Specifically,
to solve this regularizer,
we need to first group similar patches,
then denoise each patch group tensors
and finally assemble the final estimated $\mathcal{M}_i$.

However,
all these models assume the noise on $\bar{\mathcal{M}}_i$ follow univariate Gaussian distribution.
If such assumption fails,
the resulting performance can deteriorate significantly.
Here,
we have the following Proposition~\ref{pr:nslevel}.
Therefore,
the noise distribution is preserved from $\mathcal{Y}$ to $\mathcal{\bar{M}}_i$,
which enables us to use the existing spatial denoising methods.

\begin{proposition}
	\label{pr:nslevel}
	Assume the noisy HSI $\mathcal{Y}$ is from \eqref{eq:basic},
	then the noise on the reduced image $\mathcal{Y} \times_3 \mathbf{P}^{\top}$,
	where $\mathbf{P}^{\top} \mathbf{P} = \mathbf{I}$,
	still follows Gaussian distribution with zero mean and variance $\sigma_0^2$.
\end{proposition}

\begin{remark}
	While there are many other spatial denoising methods,
	e.g.,
	TV \cite{HE2016TGRS}, wavelets \cite{Chen2011TGRS,rasti2014hyperspectral} and CNN \cite{chang2018hsi},
	can be used,
	in this paper,
	we use WNNM \cite{gu2014weighted} to denoise each patch group tensor,
	as it is widely used and gives state-of-the-art denoising performance.
\end{remark}

Finally,
to use spatial denoising on each non-local group $\mathcal{G}^j$,
we need to estimate the noise level ${\sigma_i^2}$ in $\bar{\mathcal{M}}_i$,
whose noise level is changed during the iteration.
From Proposition~\ref{pr:nslevel},
we know the noisy level of $\bar{\mathcal{M}}_i$ is the same as $\mathcal{Y}_i$,
thus we propose to estimate it via
\begin{equation}
{\sigma_i=\gamma \times\sqrt{\vert \sigma_0^2 - \text{mean}( \Vert  \mathcal{Y}_i - \mathcal{Y} \Vert_F^2) \vert},}
\end{equation}
where $\gamma$ is the a scaling factor controlling the re-estimation of noise variance,
and $\text{mean}(\cdot)$ stands for the averaging process of the tensor elements.
The denoised group tensors are denoted as ${\mathcal{M}_i^j}$,
which can be directly used to reconstruct the denoised reduced image $\mathcal{M}_i$. The output denoised image of $i$-th iteration is $\mathcal{X}_i=\mathcal{M}_i\times_3\mathbf{A}_i$.

\vspace{-5px}
\subsubsection{Iterative refinement}

Iteration regularization
has been widely used to boost the denoising performance~\cite{chang2017hyper,dong2013nonlocal,gu2014weighted,xie2017kronecker}.
Here we also introduce it into our model (Algorithm \ref{alg:NSLR}) to refine the noisy orthogonal basis $\mathbf{A}_i$. As shown in \eqref{eq:diclearning}, the orthogonal basis is significantly influenced by the noise intensity of input noisy image $\mathcal{Y}_i$. Hence we update the next input noisy image as
\begin{equation*}
\mathcal{Y}_{i+1}=\lambda\mathcal{X}_i+(1-\lambda)\mathcal{Y},
\end{equation*}
where $\lambda$ is to trade-off the denoised image $\mathcal{X}_i$ and original noisy image $\mathcal{Y}$.
The estimation of $\mathbf{A}_i$ can benefit from the lower noise variance of the input $\mathcal{Y}_{i+1}$.

Besides,
$K$ is also updated with the iteration. We initialize $K$ by HySime~\cite{BioucasTGRS2008}.
When the noisy image $\mathcal{Y}$ is corrupted by heavy noise,
the estimated $K$ will be small.
Fortunately, the larger singular values obtained from the noisy image are less contaminated by the noise,
and help to keep noise out of the reduced image. With the iteration, We increase $K$ by
\begin{equation}
{K=K+\delta \times i},
\label{delta}
\end{equation}
where $\delta$ is a constant value.
Therefore, $\mathbf{A}_{i+1}$ has the ability to capture more useful information with more iterations.

\subsection{Complexity analysis}

Following the procedure of Algorithm 1, the main time complexity of each iteration includes stage A-SVD $(\mathcal{O}(MNB^2))$, stage B.non-local low-rank denoising of each $\mathcal{G}^j$ $\mathcal{O}(Tn^2K p^2)$.
Table~\ref{tab:time_com} presents the time complexity comparison between NGmeet and other non-local HSI denoising method. LLRT and KBR only need stage B to complete the denoising. As can be seen, the proposed NGmeet costs additional $\mathcal{O}(MNB^2)$ complexity in stage A,
however, will be at least $B/K$ times faster in stage B.

\begin{table}[ht]
	\footnotesize
	\vspace{-5px}
	\caption{Complexity comparison of each iteration between proposed NGmeet and state-of-the-arts non-local based methods. $\mathcal{G}^j \in  \mathbb{R}^{n\times{}n\times{}K\times{}p}$, where $n$ is the size of each patch and $p$ is the number of similar
patches. $T$ is the number of $\{\mathcal{G}^j \}$ and $T_o$ is the inner iteration of KBR.}
	\centering
	\begin{tabular}{ c | c | C{100px} }
		\hline
		       & stage A          & stage B                                       \\ \hline
		NGmeet & $\mathcal{O}(MNB^2)$ & $\mathcal{O}(T n^2Kp^2)$                             \\ \hline
		LLRT   & ---        & $\mathcal{O}(T n^2Bp^2)$                              \\ \hline
		KBR    & ---        & $\mathcal{O}(TT_0( n^2Bp(n^2+B+p)+ n^6+B^3+p^3))$     \\ \hline
	\end{tabular}
	\label{tab:time_com}
	\vspace{-10px}
\end{table}

\begin{table*}[t]
	\centering
	\footnotesize
	\caption{Quantitative comparison of different algorithms under various noise levels.
		The PSNR is in dB, and best results are in bold.}
	\begin{tabular}{c|c|c|C{27px}|C{27px}|C{27px}|C{27px}|C{27px}|C{27px}||C{27px}|C{27px}|C{27px}||C{27px}}
		\hline
		&          &       &         \multicolumn{6}{c||}{spectral low-rank methods}         & \multicolumn{3}{c||}{spatial non-local similarity methods} &                \\ \hline
		Image & $\sigma$ & Index & LRTA  & LRTV  &    MTS-NMF    & NAIL-RMA & PARA-FAC & Fast-HyDe &  TDL  &  KBR  &              LLRT              &      NGmeet      \\ \hline
		&          & PSNR  & 44.12 & 41.47 &     44.27     &  28.51   &  38.01   &   46.72   & 45.58 & 46.20 &             47.14              & \textbf{47.87} \\ \cline{3-13}
		CAVE  &    10    & SSIM  & 0.969 & 0.949 &     0.972     &  0.941   &  0.921   &   0.985   & 0.983 & 0.980 &             0.989              & \textbf{0.990} \\ \cline{3-13}
		&          &  SAM  & 7.90  & 16.54 &     8.49      &  14.52   &  13.86   &   6.62    & 6.07  & 8.94  &         \textbf{4.65}          &      4.72      \\ \cline{2-13}
		&          & PSNR  & 38.68 & 35.32 &     37.18     &  35.11   &  37.58   &   41.21   & 39.67 & 41.52 &             42.53              & \textbf{43.11} \\ \cline{3-13}
		&    30    & SSIM  & 0.913 & 0.818 &     0.855     &  0.775   &  0.888   &   0.945   & 0.942 & 0.942 &         \textbf{0.974}         &     0.972      \\ \cline{3-13}
		&          &  SAM  & 12.86 & 33.32 &     14.97     &  32.43   &  17.37   &   14.06   & 12.54 & 19.43 &              8.23              & \textbf{7.46}  \\ \cline{2-13}
		&          & PSNR  & 35.49 & 32.27 &     33.40     &  32.11   &  30.06   &   38.05   & 36.51 & 39.41 &             40.09              & \textbf{40.45} \\ \cline{3-13}
		&    50    & SSIM  & 0.858 & 0.719 &     0.730     &  0.638   &  0.571   &   0.889   & 0.888 & 0.922 &             0.950              & \textbf{0.951} \\ \cline{3-13}
		&          &  SAM  & 16.53 & 43.65 &     19.06     &  22.85   &  38.35   &   20.08   & 18.23 & 21.31 &             11.48              & \textbf{9.80}  \\ \cline{2-13}
		&          & PSNR  & 31.21 & 27.97 &     27.96     &  27.90   &  24.29   &   33.41   & 31.90 & 33.78 &             36.25              & \textbf{37.21} \\ \cline{3-13}
		&   100    & SSIM  & 0.735 & 0.529 &     0.493     &  0.453   &  0.256   &   0.746   & 0.734 & 0.851 &             0.910              & \textbf{0.927} \\ \cline{3-13}
		&          &  SAM  & 22.67 & 54.85 &     26.33     &  55.66   &  51.83   &   30.72   & 28.51 & 26.41 &             18.17              & \textbf{16.23} \\ \hline
		&          & PSNR  & 38.49 & 38.71 &     40.64     &  41.46   &  33.39   &  42.220   & 41.46 & 40.09 &             41.95              & \textbf{43.17} \\ \cline{3-13}
		PaC  &    10    & SSIM  & 0.975 & 0.979 &     0.988     &  0.987   &  0.866   &   0.990   & 0.988 & 0.984 &             0.989              & \textbf{0.992} \\ \cline{3-13}
		&          &  SAM  & 4.90  & 3.29  &     2.76      &   3.46   &   9.05   &   2.99    & 3.06  & 2.86  &              2.75              & \textbf{2.61}  \\ \cline{2-13}
		&          & PSNR  & 32.07 & 32.76 &     35.45     &  34.17   &  30.92   &   35.98   & 34.43 & 34.39 &             35.04              & \textbf{36.97} \\ \cline{3-13}
		&    30    & SSIM  & 0.908 & 0.920 &     0.958     &  0.941   &  0.845   &   0.962   & 0.949 & 0.947 &             0.957              & \textbf{0.971} \\ \cline{3-13}
		&          &  SAM  & 7.88  & 5.76  & \textbf{4.17} &   6.54   &   9.28   &   5.09    & 5.11  & 4.28  &              4.86              &      4.30      \\ \cline{2-13}
		&          & PSNR  & 29.11 & 29.45 &     32.51     &  30.71   &  29.24   &   33.32   & 31.31 & 31.05 &             32.00              & \textbf{34.29} \\ \cline{3-13}
		&    50    & SSIM  & 0.836 & 0.850 &     0.921     &  0.886   &  0.846   &   0.936   & 0.904 & 0.892 &             0.918              & \textbf{0.948} \\ \cline{3-13}
		&          &  SAM  & 9.20  & 8.60  &     5.50      &   8.83   &  11.40   &   6.55    & 6.14  & 5.40  &              6.55              & \textbf{5.18}  \\ \cline{2-13}
		&          & PSNR  & 25.13 & 26.22 &     28.17     &  25.76   &  23.68   &   29.90   & 27.49 & 27.80 &             28.63              & \textbf{30.61} \\ \cline{3-13}
		&   100    & SSIM  & 0.655 & 0.729 &     0.808     &  0.728   &  0.598   &   0.873   & 0.789 & 0.793 &             0.833              & \textbf{0.890} \\ \cline{3-13}
		&          &  SAM  & 10.17 & 12.76 &     8.40      &  12.93   &  20.22   &   8.68    & 7.67  & 6.95  &              7.68              & \textbf{6.86}  \\ \hline
		&          & PSNR  & 38.94 & 36.64 &     37.26     &  42.57   &  32.38   &   43.06   & 41.83 & 40.58 &             41.89              & \textbf{43.72} \\ \cline{3-13}
		WDC  &    10    & SSIM  & 0.974 & 0.968 &     0.975     &  0.989   &  0.914   &   0.991   & 0.989 & 0.986 &             0.990              & \textbf{0.993} \\ \cline{3-13}
		&          &  SAM  & 5.602 & 4.653 &     4.429     &  3.637   &  8.087   &   3.070   & 3.680 & 3.090 &             3.700              & \textbf{2.830} \\ \cline{2-13}
		&          & PSNR  & 32.91 & 32.42 &     34.65     &  35.87   &  31.56   &  37.390   & 34.84 & 34.75 &             36.30              & \textbf{37.90} \\ \cline{3-13}
		&    30    & SSIM  & 0.917 & 0.909 &     0.953     &  0.958   &  0.898   &   0.971   & 0.953 & 0.951 &             0.967              & \textbf{0.975} \\ \cline{3-13}
		&          &  SAM  & 8.331 & 5.991 &     5.557     &  7.011   &  9.009   &   5.140   & 6.400 & 5.240 &             5.460              & \textbf{4.640} \\ \cline{2-13}
		&          & PSNR  & 30.35 & 30.12 &     32.49     &  32.56   &  29.49   &   34.61   & 31.89 & 31.61 &             33.48              & \textbf{35.14} \\ \cline{3-13}
		&    50    & SSIM  & 0.864 & 0.849 &     0.922     &  0.919   &  0.837   &   0.948   & 0.910 & 0.900 &             0.938              & \textbf{0.955} \\ \cline{3-13}
		&          &  SAM  & 9.43  & 7.09  &     6.71      &   9.22   &  13.64   &   6.57    & 7.94  & 6.63  &              6.43              & \textbf{5.83}  \\ \cline{2-13}
		&          & PSNR  & 26.84 & 27.23 &     28.94     &  27.85   &  23.01   &   31.05   & 27.66 & 28.23 &             29.88              & \textbf{31.45} \\ \cline{3-13}
		&   100    & SSIM  & 0.734 & 0.740 &     0.830     &  0.805   &  0.550   &   0.894   & 0.781 & 0.789 &             0.861              & \textbf{0.903} \\ \cline{3-13}
		&          &  SAM  & 11.33 & 9.47  &     9.44      &  13.27   &  25.46   &   8.91    & 10.15 & 9.12  &              7.99              & \textbf{7.86}  \\ \hline
	\end{tabular}%
	\label{tab:simucase}%
\end{table*}%

\section{Experiments}
In this section, we present the simulated and real data experimental results of different methods, companied with the computational efficiency and parameter analysis of the proposed NGmeet. The experiments are programmed in Matlab with CPU Core i7-7820HK 64G memory.

\subsection{Simulated experiments}

\noindent
\textbf{Setup.}
One multi-spectral image (MSI) CAVE
\footnote{\url{http://www1.cs.columbia.edu/CAVE/databases/}}, and two HSI images,
\textit{i.e.}
PaC
\footnote{\url{http://www.ehu.eus/ccwintco/index.php/}}
and WDC
\footnote{\url{https://engineering.purdue.edu/~biehl/MultiSpec/hyperspectral}}
datasets are used (Table~\ref{tab:imagesize}).
These images have been widely used for a simulated study
\cite{chang2017hyper,he2015hyperspectral,CVPR2014Meng,xie2017kronecker,zhuang2018fast}.
Following the settings in~\cite{chang2017hyper,CVPR2014Meng},
additive Gaussion noise with noise variance $\sigma_0^2$ are added to the MSIs/HSIs with $\sigma_0^2$ varies from $10, 30, 50$ to $100$. Before denoising, the whole HSIs are normalized to [0, 255].

\begin{table}[ht]
	\footnotesize
	\caption{Hyper-spectral images used for simulated experiments.}
	\centering
	\begin{tabular}{ c | c | c | c }
		\hline
		& CAVE           & PaC            & WDC            \\ \hline
		image size     & 512$\times$512 & 256$\times$256 & 256$\times$256 \\ \hline
		number of bands & 31             & 89             & 192            \\ \hline
	\end{tabular}
	\label{tab:imagesize}
	\vspace{-7px}
\end{table}

\begin{figure*}[t]
	\centering
	\includegraphics[width=0.85\linewidth]{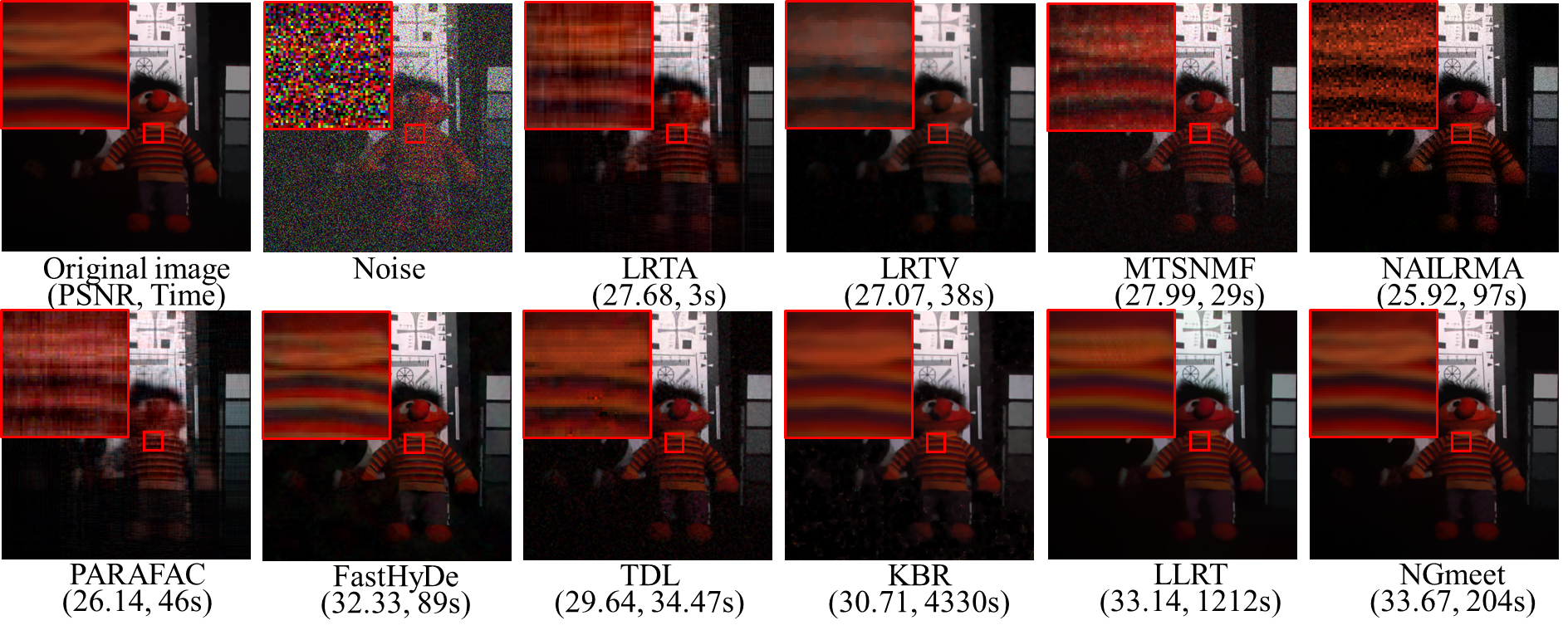}
	\vspace{-5px}
	\caption{Denoising results on the CAVE-toy image with the noise variance 100. The color image is composed of bands 31, 11, and 6 for the red, green, and blue channels, respectively.}
	\label{fig:simu1}
	\vspace{-10px}
\end{figure*}

The following methods are used for the comparison:
\textit{spectral low-rank methods}, \textit{i.e.}
LRTA~\cite{renard2008denoising}
\footnote{\url{https://www.sandia.gov/tgkolda/TensorToolbox/}},
LRTV~\cite{HE2016TGRS}
\footnote{\url{https://sites.google.com/site/rshewei/home}},
MTSNMF~\cite{QianyuntaoTGRS2015}
\footnote{\url{http://www.cs.zju.edu.cn/people/qianyt/}},
NAILRMA~\cite{he2015hyperspectral}
PARAFAC~\cite{liu2012denoising}
and FastHyDe~\cite{zhuang2018fast}
\footnote{\url{http://www.lx.it.pt/~bioucas/}};
\textit{spatial non-local similarity methods}, \textit{i.e.}
TDL~\cite{CVPR2014Meng}
KBR~\cite{xie2017kronecker}
\footnote{\url{http://gr.xjtu.edu.cn/web/dymeng/}},
LLRT~\cite{chang2017hyper}
\footnote{\url{http://www.escience.cn/people/changyi/}};
and finally
NGmeet\footnote{\url{https://github.com/quanmingyao/NGMeet}} (Algorithm~\ref{alg:NSLR}),
which combines the best of above two fields.
Hyper-parameters of all compared methods are set based on authors' codes or suggestions in the paper.
The value of spectral dimension $K$ is the most import parameter, which is initialized by HySime~\cite{BioucasTGRS2008} and updated via \eqref{delta}.
Parameter $\mu$ is used to control the contribution of non-local regularization, and $\gamma$ is a scaling factor controlling the re-estimation of noise variance~\cite{dong2013nonlocal}.
We empirically set $\mu=1$, $\lambda = 0.9$ and $\gamma = 0.5$ as introduced in~\cite{chang2017hyper}, and $\delta = 2$ in the whole experiments.

To thoroughly evaluate the performance of different methods, the peak signal-to-noise ratio (PSNR) index,
the structural similarity (SSIM)~\cite{SSIM2004image} index and the
spectral angle mean (SAM)~\cite{chang2017hyper,HE2016TGRS} index were adopted to give a quantitative assessment.
The SAM index is to measure the mean spectrum degree between the original HSI and the restored HSI.
The lower value of SAM means the higher similarity between original image and the denoised image.

\noindent
\textbf{Quantitative comparison.}
For each noise level setting, we calculate  evaluation values of all the images from each dataset, as presented in Table \ref{tab:simucase}. It can be easily observed that the proposed NGmeet method achieved the best results almost in all cases. Another interesting observation is that the non-local based method LLRT can achieve better results than FastHyDe, the best result of spectral low-rank methods, but it dose the opposite in the hyperspectral image cases. This phenomenon conforms the advantage of NL low-rank property in the MSI processing and the spectral low-rank property in the HSI processing.

\noindent
\textbf{Visual comparison.}
To further demonstrate the efficiency of the proposed method,
Figure~\ref{fig:simu1} shows the color images of CAVE-toy (composed of bands 31, 11 and 6~\cite{he2015hyperspectral}) before and after denoising. The results of PaC and WDC can be found in the supplementary material. The PSNR values and the computational time of each methods are marked under the denoised images. It can be observed that FastHyDe, LLRT and NGmeet have huge advantage over the rest comparison methods. From the enlarged area, the results of FastHyDe LLRT produced some artifacts.
Thus, our method NGmeet can produce the best visual quality.

 \vspace{3px}
\noindent
\textbf{Computational efficiency.}
In this section, we will illustrate that in our denoising paradigm, the computational efficiency of the non-local denoising procedure will get rid of the huge spectral dimension.
Compared to the previous non-local denoising methods, \textit{i.e.} KBR~\cite{xie2017kronecker} and LLRT~\cite{chang2017hyper}, the proposed NGmeet includes additional stage A.
Table~\ref{tab:time} presents the computational time of different stages of the three methods. From Table~\ref{tab:time_com} and~\ref{tab:time}, we can conclude that
NGmeet spends little time to project the original HSI onto a reduced image (stage A), however, earning huge advantage in stage B including group matching step and non-local denoising.

\begin{table}[ht]
	\centering
	\vspace{-5px}
	\caption{Average running time (in seconds) of each stage for the non-local low-rank based methods.
		stage A: spectral low-rank denoising;
		stage B: spatial non-local low-rank denoising.}
	\footnotesize
	\begin{tabular}{c|c|c|c|c|c}
		\hline
		  Time    & KBR  & LLRT &      \multicolumn{3}{c}{NGmeet}       \\ \cline{2-6}
		(seconds) & stage B & stage B & stage A  & stage B & total \\ \hline
		  CAVE    & 4330  & 1212  &    3     &  201    &  204 \\ \hline
		   PaC    & 828   & 488   &    2     &  37     &  39   \\ \hline
		   WDC    & 3570  &1573   &    3     &  45     &  48   \\ \hline
	\end{tabular}
	\label{tab:time}
	\vspace{-5px}
\end{table}%

\begin{figure}
  \centering
  \subfigure[Time v.s. number of bands]{
    \includegraphics[width= 0.45 \linewidth]{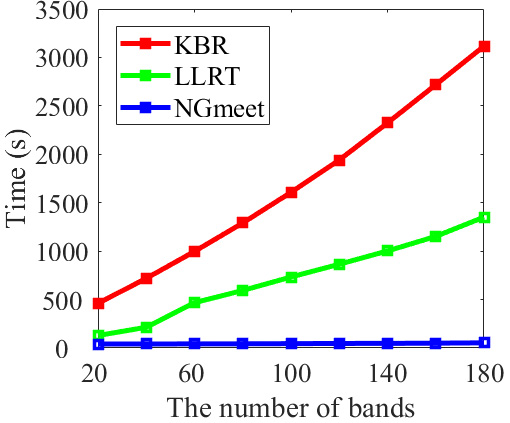}
  }
  \subfigure[SSIM v.s. number of bands]{
    \includegraphics[width= 0.45 \linewidth]{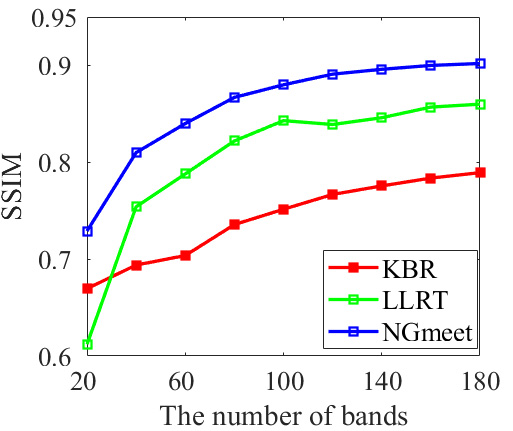}
  }
  \caption{The computational time and SSIM values of different numbers of bands.
  	WDC is used and noise variance is 100.}
  \label{fig:ILL1}
  \vspace{-10px}
\end{figure}

Figure~\ref{fig:ILL1} displays the computational time and SSIM values of the proposed NGmeet, KBR~\cite{xie2017kronecker} and LLRT~\cite{chang2017hyper}, with the increase of spectral number. As illustrated, even though the performances of KBR and LLRT increase with the increase of spectral number, the computational time also increases linearly. Our method can achieve the best performance, meanwhile, the computational time is nearly unchanged with the increase of spectral number.

\begin{figure}[ht]
	\centering
	\vspace{-10px}
	\includegraphics[width=0.5\linewidth]{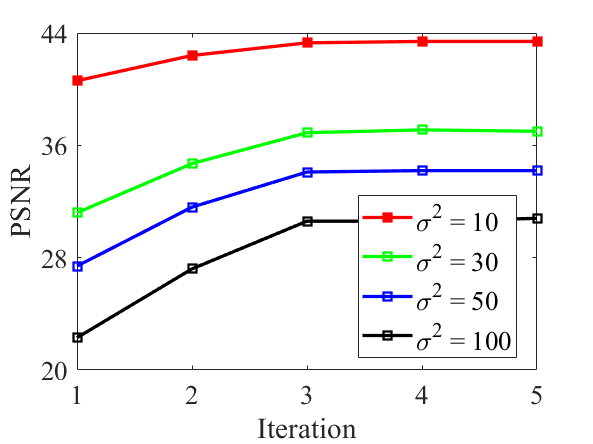}
	\vspace{-5px}
	\caption{PSNR v.s. iteration of NGmeet. WDC is used.}
	\label{fig:PSNRvsite}
	\vspace{-10px}
\end{figure}

\noindent
\textbf{Convergence.}
To show the convergence of proposed NGmeet,
Figure~\ref{fig:PSNRvsite} presents the PSNR values with the increase of iteration, on the WDC dataset.
From the figure, it can be observed that our method can converge to stable PSNR values very fast
at different noise level.

\subsection{Real Data Experiments}
\begin{figure*}[ht]
\centering
\includegraphics[width=0.85\linewidth]{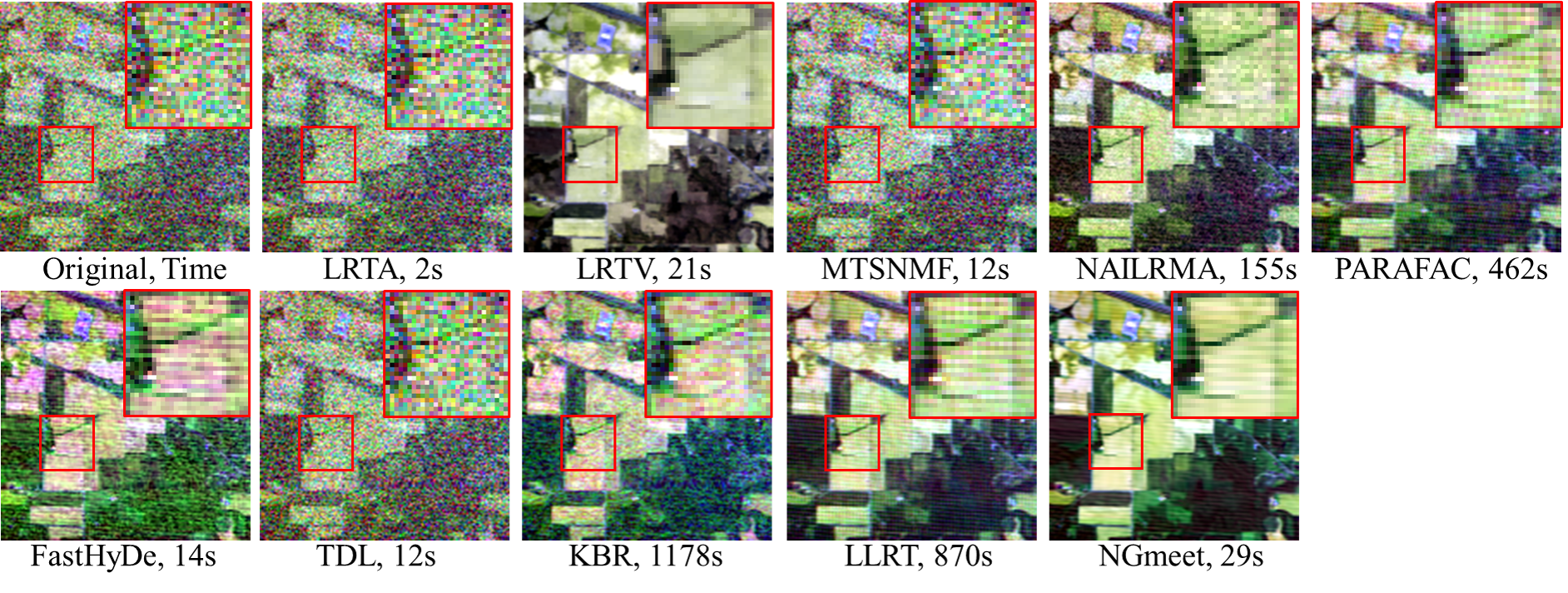}
\vspace{-5px}
\caption{Real data experimental results on the Indian Pines dataset. The color image is composed of noisy bands 219, 109 and 1.}
\label{fig:realindian}
\vspace{-5px}
\end{figure*}

\begin{figure*}[ht]
\centering
\includegraphics[width=0.85\linewidth]{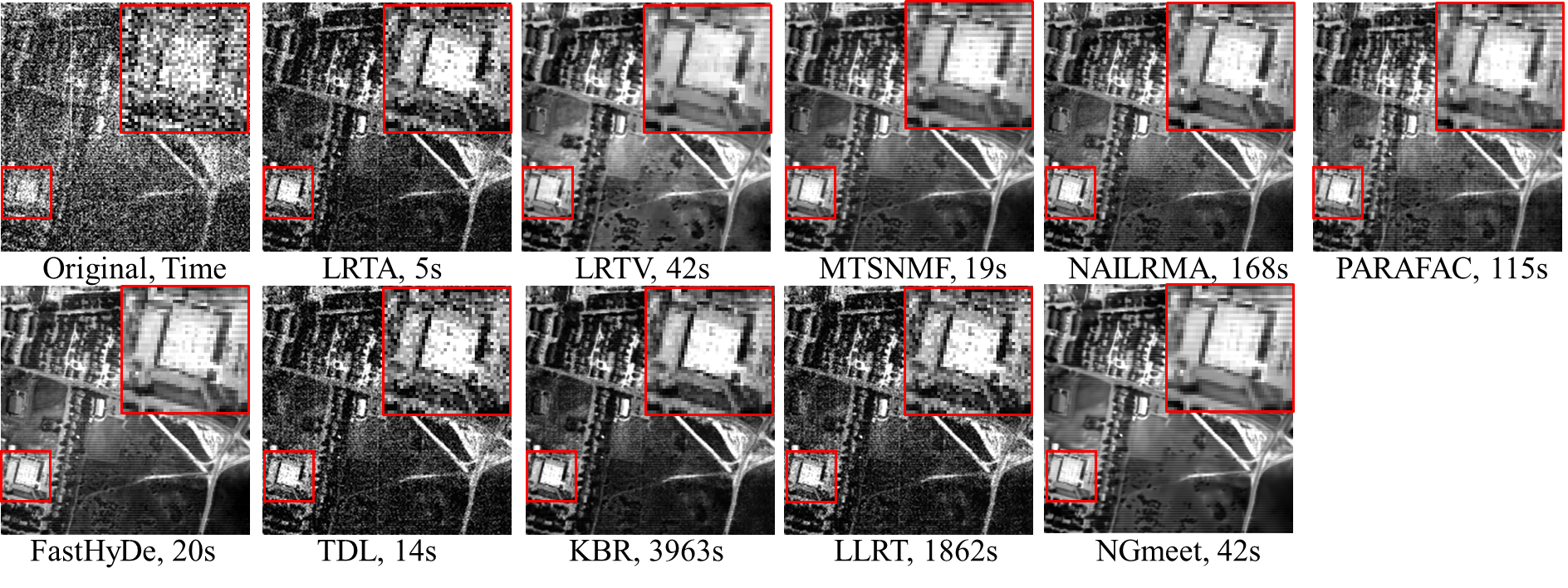}
\vspace{-5px}
\caption{Real data experimental results on the Urban dataset of band 207.}
\label{fig:realUrban}
\vspace{-15px}
\end{figure*}

\noindent
\textbf{Setup.}
Here, AVIRIS Indian Pines HSI
\footnote{\url{https://engineering.purdue.edu/~biehl/MultiSpec/}}
and HYDICE Urban image
\footnote{\url{http://www.tec.army.mil/hypercube}}
are adopted in the real experiments (Table~\ref{tab:realdata}). As in \cite{He2014TGRS}, 20 water absorption bands (104-108, 150-163, 220 bands) of Indian Pines are excluded for illustration, since they do not contain useful information.
The noisy HSIs are also scaled to the range [0 255], and the parameters involved in the proposed methods are set as the same in the simulated experiments.
In addition, multiple regression theory-based approach~\cite{BioucasTGRS2008} is adopted to estimate the initial noise variance of each HSI bands.

\begin{table}[ht]
\footnotesize
\centering
\vspace{-5px}
\caption{Hyperspectral images used for real data experiments.}
\begin{tabular}{c | c | c}
	\hline
	               & Urban          & Indian Pines        \\ \hline
	  image size   & 200$\times$200 & 145$\times$145 \\ \hline
	number of bands & 210            & 220            \\ \hline
\end{tabular}
\label{tab:realdata}
\vspace{-5px}
\end{table}

\noindent
\textbf{Visual comparison.}
Since reference clean images are missing, we just present the real Indian Pines and Urban images before and after denoising in figures~\ref{fig:realindian} and~\ref{fig:realUrban}. It can be obviously observed that the results produced by the proposed NGmeet can remove the noise and keep the spectral details simultaneously. LRTV can produce the most smooth results. However, the color of the denoised result changes a lot, indicating the loss of spectral information. The denoised results of FastHyDe and LLRT still contain stripes as presented in Figure~\ref{fig:realindian}. To sum up, although the proposed NGmeet is designed in the Gaussion noise assumption, it can also achieves the best results for real datasets.

\subsection{Parameter analysis}

\begin{figure}[ht]
	\centering
	\includegraphics[width = 0.5 \linewidth]{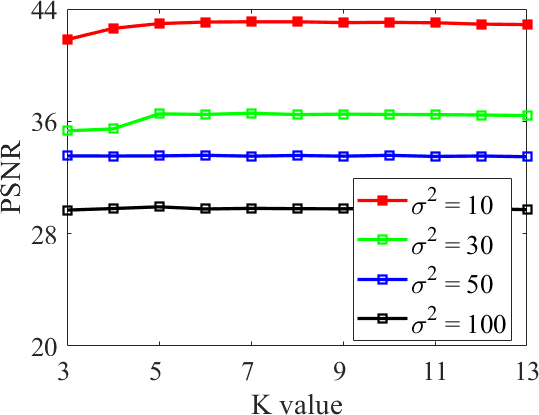}
	\caption{PSNR values achieved by the proposed methods with different parameter $K$ with $\delta=0$ on the PaC dataset.}
	\label{fig:K_analysis}
\end{figure}

$K$ is the key parameter to integrate the spatial and spectral information. Figure~\ref{fig:K_analysis} presents the PSNR values achieved by NGmeet with different initialization of $K$ with
$\delta$ being $0$. PaC images was chosen as the test image, and the
noise variance $\sigma_0^2$ changes from $10, 30, 50$ to $100$. $K$ is initialized by HySime~\cite{BioucasTGRS2008} as $7, 6, 6, 5$ for different noise variance cases, respectively. It confirms that the initialization of $K$ is reliable.

Table~\ref{tab:delta_analysis} presents the influence of different $\sigma_0^2$ values with $K$ being initialized by HySime~\cite{BioucasTGRS2008}.
It can be observed that, the updating strategy of $K$ can improve the performance, and the selection of $\delta$ is robust.

\begin{table}[ht]
\centering
\footnotesize
\vspace{-5px}
\caption{The influence of different $\delta$ for NGmeet.}
\begin{tabular}{c|c|c|c|c}
	\hline
	  PSNR(dB)   & $\sigma_0^2=10$ & $\sigma_0^2=30$ & $\sigma_0^2=50$ & $\sigma_0^2=100$ \\ \hline
	$\delta=0 $  &    43.09    &    36.49    &    33.54    &    29.91     \\ \hline
	$ \delta=1 $ &    43.52    &    36.96    &    34.23    &    30.56     \\ \hline
	$ \delta=2 $ &    43.43    &    37.02    &    34.21    &    30.83     \\ \hline
	$ \delta=3 $ &    43.42    &    37.11    &    34.42    &    30.45     \\ \hline
\end{tabular}%
\label{tab:delta_analysis}
\vspace{-10px}
\end{table}%

\section{Conclusion}

In this paper,
we provide a new perspective to integrate the spatial non-local similarity and global spectral low-rank property, which are explored by low-dimensional orthogonal basis and reduced image denoising, respectively. We have also proposed an alternating minimization method with iteration strategy to solve the optimization of the proposed GNmeet method. The superiority of our method are confirmed by the simulated and real dataset experiments.
In our unified spatial-spectral paradigm,
the usage of WNNM \cite{gu2014weighted} is not a must.
In future,
we plan to adopt Convolutional Neural Network~\cite{chang2018hsi,zhang2017learning,yuanHSID-CNN2018} to explore non-local similarity;
and automated machine learning \cite{yao2018taking} to help tuning and configuring hyper-parameters.


\cleardoublepage
{
\small
\bibliographystyle{ieee}
\bibliography{lowrank_review_references}
}

\cleardoublepage
\section{Appendix}

\subsection{Proof}
\subsubsection{Proposition 3.1}
\begin{proof}
Note that the objective can be expressed as:
\begin{align*}
\min_{\mathbf{M}_{(3)},\mathbf{A}^{\top} \mathbf{A}=\mathbf{I}}
\frac{1}{2}
\| (\mathbf{Y}_i)_{(3)} - \mathbf{M}_{(3)} \mathbf{A} \|_F^2,
\end{align*}
which is equal to find the best $K$-rand approximation of $(\mathbf{Y}_i)_{(3)}$.
Thus,
let rank-$K$ SVD of $(\mathbf{Y}_i)_{(3)}$ be $\mathbf{U}$, $\mathbf{S}$ and $\mathbf{V}$,
the closed-form solution of (4) in the paper
is given by $\mathbf{A}_i = \mathbf{V}$ and $\bar{\mathcal{M}_i} = \text{\normalfont fold}_3(\mathbf{U} \mathbf{S})$.
\end{proof}

\subsubsection{Proposition~3.2}

\begin{proof}
Since $\mathcal{Y} = \mathcal{X} + \mathcal{N}$,
then
\begin{align}
\mathcal{Y} \times_3 \mathbf{P} = \mathcal{X} \times_3 \mathbf{P} + \mathcal{N} \times_3 \mathbf{P},
\end{align}
where the noise is given by  $\mathcal{N} \times_3 \mathbf{P}$.
Note that
\begin{equation}
\text{mean}
\left[ \mathcal{N} \times_3 \mathbf{P} \right]  = \mathbf{0}.
\end{equation}
Thus, the mean of the noise is zero.
Let $\mathbf{a}$ be a column in $\mathbf{N}_{(3)}$,
then
one column $\mathbf{b}$ in $(\mathcal{N} \times_{3} \mathbf{P})_{(3)}$ can be expressed as
\begin{equation}
\mathbf{b} = \mathbf{P} \mathbf{a}.
\end{equation}
Follow the definition of variance,
we have
\begin{align*}
\text{var}\left[ \mathbf{b} \right]
& = \text{mean}
\left[ (\mathbf{b} - \text{mean} \left[ \mathbf{b} \right])^2  \right]
\\
& = \text{mean}
\left[ \mathbf{b} \mathbf{b}^{\top} \right]
\\
& = \text{mean}\left[ \mathbf{a}^{\top} \mathbf{P}^{\top} \mathbf{P} \mathbf{a}  \right]
\\
& = \text{mean}\left[ \mathbf{a}^{\top} \mathbf{a}  \right]  = \sigma_0 \mathbf{I}.
\end{align*}
Thus,
we obtain the proposition.
\end{proof}

\subsection{Extra Experiments Results}

\begin{figure*}[!ht]
	\centering
	\includegraphics[width=0.85\linewidth]{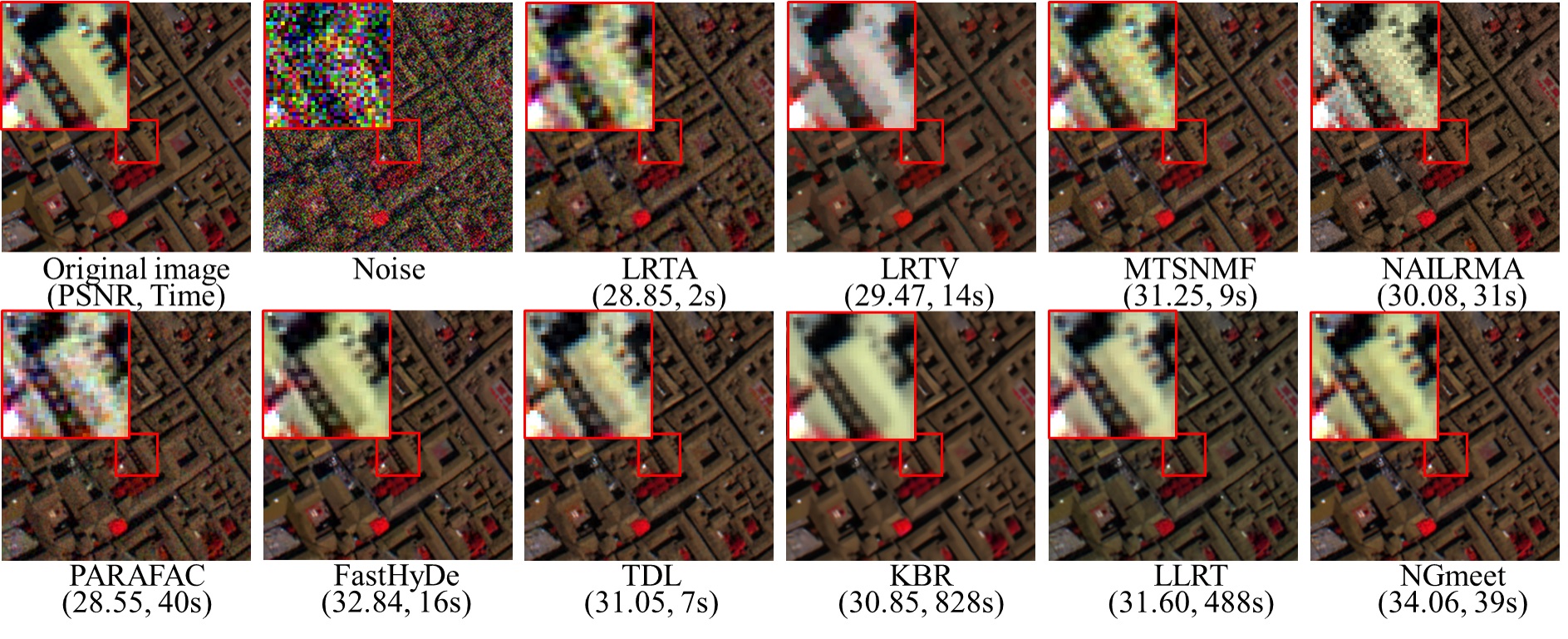}
	\vspace{-5px}
	\caption{Denoising results on the PaU image with the noise variance 50. The color image is composed of bands 80, 34, and 9 for the red, green, and blue channels, respectively.}
	\label{fig:simu2}
\end{figure*}

\begin{figure*}[!ht]
	\centering
	\includegraphics[width=0.85\linewidth]{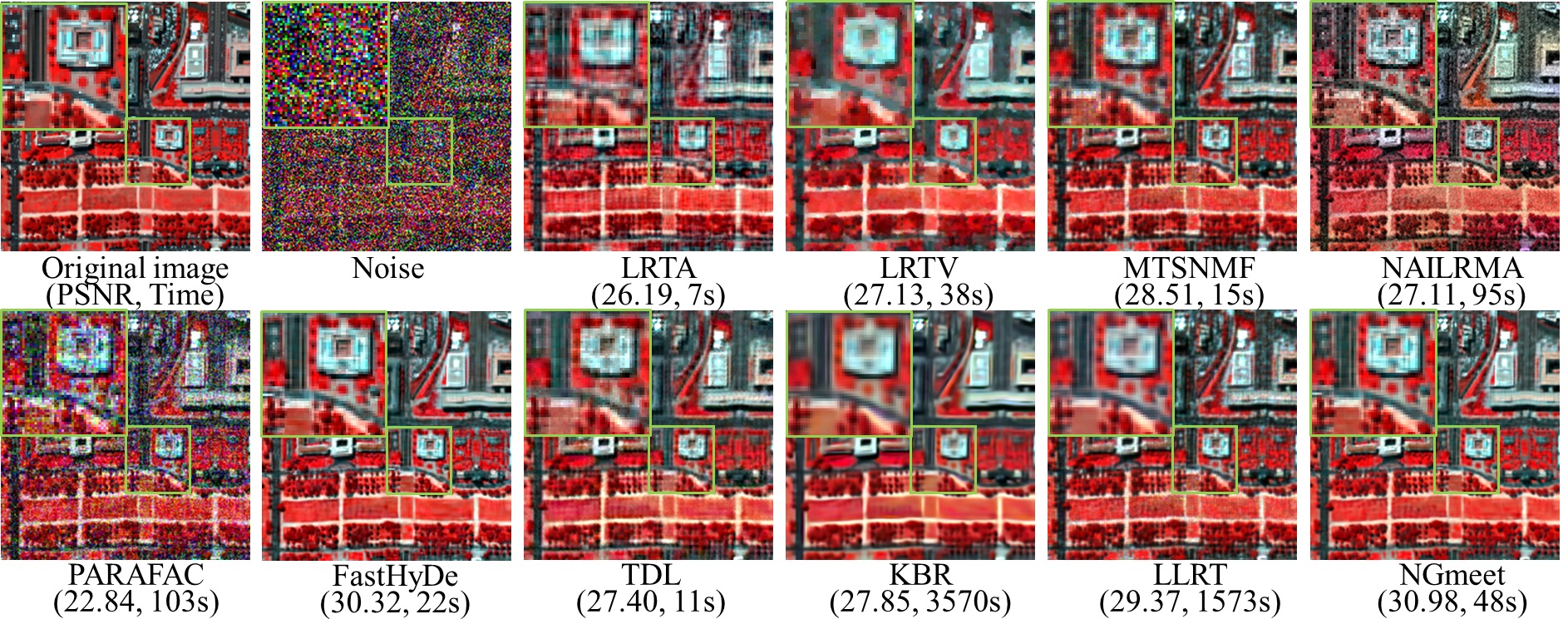}
	\vspace{-5px}
	\caption{Denoising results on the WDC image with the noise variance 100. The color image is composed of bands 190, 60 and 27 for the red, green, and blue channels, respectively.}
	\label{fig:simu3}
\end{figure*}

Figure~\ref{fig:simu2} and ~\ref{fig:simu3} show the color images of PaU~\cite{he2015hyperspectral} (composed of bands 80, 34 and 9) and WDC~\cite{yuanHSID-CNN2018} (composed of bands 190, 60 and 27) before and after denoising.

\end{document}